\documentclass[10pt,twocolumn,letterpaper]{article}
\usepackage{times}
\usepackage{epsfig}
\usepackage{graphicx}
\usepackage{amsmath}
\usepackage{amssymb}
\usepackage{color}

\begin{document}

\title{Stochastic Trajectory Prediction with Social Graph Network}
\date{}

\author{Lidan Zhang\\
Intel Labs China\\
{\tt\small lidan.zhang@intel.com}
\and
Qi She\\
Intel Labs China\\
{\tt\small qi.she@intel.com}
\and
Ping Guo\\
Intel Labs China\\
{\tt \small ping.guo@intel.com}
}

\maketitle

\newcommand{\etal}{\textit{et al.}}



\begin{abstract}
    Pedestrian trajectory prediction is a challenging task because of the complexity of real-world human social behaviors and uncertainty of the future motion. For the first issue, existing methods adopt fully connected topology for modeling the social behaviors, while ignoring non-symmetric pairwise relationships. To effectively capture social behaviors of relevant pedestrians, we utilize a directed social graph which is dynamically constructed on timely location and speed direction. Based on the social graph, we further propose a network to collect social effects and accumulate with individual representation, in order to generate destination-oriented and social-aware representations. For the second issue, instead of modeling the uncertainty of the entire future as a whole, we utilize a temporal stochastic method for sequentially learning a prior model of uncertainty during social interactions. The prediction on the next step is then generated by sampling on the prior model and progressively decoding with a hierarchical LSTMs. Experimental results on two public datasets show the effectiveness of our method, especially when predicting trajectories in very crowded scenes.
\end{abstract}

\section{Introduction}
\label{sec:intro}
Given an observation of history trajectory, learning to predict future pedestrian locations is an essential task in many applications, e.g. autonomous driving, robot navigation, etc. Although tremendous research have been investigated~\cite{Alahi16cvpr,Becker2018,Lee17cvpr}, it is still challenging to capture the complex social interactions in crowded scenarios. For example, a person can walk alone or with others in a consistent group. A group might be changed when a person joins or leaves. As a result, the individual trajectory is usually influenced by others in order to avoid collisions while following reasonable social norms. Furthermore, the future routine is always ambiguous, meaning that more than one path are reasonable to reach the same destination.

Motivated by these, we highlight two factors that are crucial in the trajectory prediction:

\textit{1). Social interaction between human is non-symmetric.}

Compared with previous works~\cite{Alahi16cvpr,Gupta18cvpr,Vemula18icra,Zhang19cvpr}, we propose that the pairwise social interaction should be non-symmetric among pedestrians. For example, a person always pays attention to pedestrians ahead of him, and has little aware of pedestrians behind him. To simulate this, we model the social topology as a directed graph. Then we propose a graph network which accumulates social interactions to the intrinsic destination in order to capture destination-oriented feature enriched with social interaction patterns.

\textit{2). The next step is uncertain, dependent on both intrinsic destination and social selection at each time.}

During walking, a person might adopt various flexible decisions in avoid collision. To model this uncertainty, previous work introduced one single random variable sampled from either the prior distribution conditioned on the observations \cite{Lee17cvpr} or a fixed multivariate normal distribution \cite{Gupta18cvpr,Sadeghian18}. The single random variable will be used to generate all future steps. However, in the real-world, the selection on the next step might be changed during walking. For example, a person might try to surpass the person ahead of him at first, but he might give up and continue to follow. To model the temporal stochastic and generate diverse routines, at each time step, one latent variable is sampled from a learned distribution, which models all possible selections on the next step till the moment.

In summary, in this paper, we propose two contributions to generate all destination-oriented and social-plausible future trajectories. A new social graph network can effectively extract non-symmetric pairwise relationships and social interactions. A stochastic method can predict diverse social-plausible selections for the next step. The final stochastic predictions are generated by progressively integrating both social and individual destination information with a  hierarchical LSTM.

\section{Related Work}
\label{sec:related}
Predicting the future is always a challenging problem in computer vision. It has been widely investigated in many fields, such as video frame prediction (\cite{Babaeizadeh18iclr,Emily18icml}), motion flow prediction \cite{walker2016eccv}, traffic forecasting\cite{li18iclr}, car trajectory prediction\cite{Rhinehart18eccv}, etc. For predicting pedestrian trajectories, a lot of approaches have been proposed to make predictions for the first person view ~\cite{park16cvpr,Yagi18cvpr}, collaboration with non-homogeneous traffic agents~\cite{Yexin19AAAI}, team sports\cite{Felsen18eccv}, etc. In this paper, we focus on prediction under the fixed camera view given only world coordinates inputs.

\noindent\textbf{Trajectory Prediction:} The earlier works used heuristic features to model human-human interactions. For example, in social force model ~\cite{Helbing95}, each trajectory is generated by applying both attractive forces towards an intended destination and repulsive forces to avoid collisions. But the social forces only consider individual history, Linear Trajectory Avoidance (LTA ~\cite{Pellegrini09iccv}) predicts the future trajectory by jointly anticipating the movement of other pedestrians and obstacles in the scene.

Over the past several years, data-driven methods based on RNNs have showed powerful ability in modeling sequential data. Based on RNN, Social LSTM~\cite{Alahi16cvpr} firstly proposed social pooling, which updates the hidden state of the each pedestrian by summing up the states of neighborhood pedestrians with predefined regular grids. In order to remove the limitation of neighborhood grids, the social pooling is extended to a multi-layer perceptron (MLP) network in ~\cite{Gupta18cvpr}.  SR-LSTM~\cite{Zhang19cvpr} iteratively refines the cell and hidden states in LSTM at each time-step by learning attention on other pedestrians. Similar attention mechanism can also be found in ~\cite{Vemula18icra}. CIDNN~\cite{Xu18cvpr} used spatial affinity to replaced attention weights, which are the inner product on the embedding representations of the current locations. To improve training efficiency, adversarial training is also introduced in ~\cite{Gupta18cvpr,Zou18aaai}.

\noindent\textbf{Stochastic Prediction:} In trajectory prediction, most of the previous works use a single stochastic variable to model possible diversity. Based on conditional auto-encoder (CVAE,~\cite{kihyuk15nips}) framework, Lee \etal samples latent variables conditioning on the summary of the observed trajectory given by RNN, then decodes into a sequence. However, they did not consider pedestrian interactions during generation. SoPhie~\cite{Sadeghian18} modified decoder LSTM by sampling a white noise and concatenated with the scene information extracted by CNN as inputs. Su \etal ~\cite{Su17ijcai} adds $2$ Gaussian processes on LSTM hidden states, in order to generate probabilistic predictions. Similarly, a stochastic extension of LTA~\cite{Pellegrini10sism} obtains a set of possible future states by extending the original energy into a probabilistic form, then estimate the Gibbs potential by fitting a Gaussian mixture model.

VRNN~\cite{vrnn15nips} firstly introduces temporal stochastic latent variable, which is dependent on both the current LSTM hidden state and the input at each time step. Based on VRNN, \cite{Fraccaro16nips,Anirudh17nips} extend the stochastic latent variables to be time-dependent and adds a auxiliary backward LSTM for training. The key difference of these models are the choice of prior, approximated posterior model and loss function. A recent work in~\cite{Chen19iclr} is similar to our work, which associates hidden states in VRNN with a fully connected graph interaction network. Different from us, they introduce context image as additional input, and generate predictions as a weighted combination of visual decoder and VRNN decoder. Furthermore, they studied team sports with highly collaborative agents, whereas the social interactions in our work is more flexible, with both collaborative and standalone agents. Our work is also inspired by \cite{Emily18icml} on learning dynamic prior model across time. The difference is that we use the output of social graph for modeling uncertainty and propose a hierarchical LSTMs as decoder to progressively integrate different information.  Although the stochastic framework is similar, \cite{Emily18icml} aims to model frame uncertainty in video prediction, whereas our goal is to capture social interaction uncertainty in trajectory prediction problem.

\begin{figure*}[!htp]
 \centering
 \includegraphics[width=.95\linewidth]{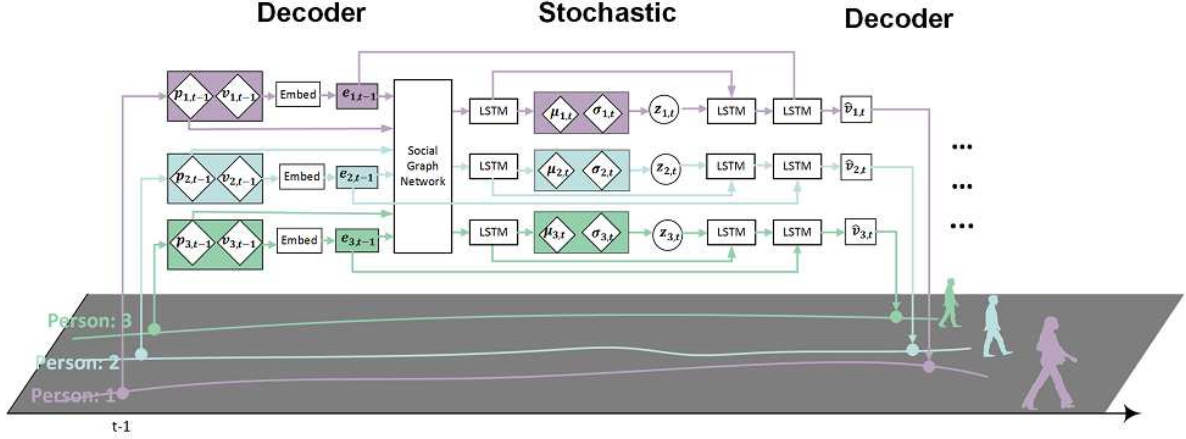}
 \vspace{-5mm}
  \caption{Illustration of the overall approach. At each time-step, each agents' status(location and velocity) is processed through social graph network and LSTM to encode both individual and social interactions. Then a stochastic variable is sampled from a prior Gaussian distribution, and finally used to generate predictions on agents' movement. We use a diamond shape to represent deterministic variable, and use a circle to represent stochastic variable. For brevity, the inference model is omitted.}
  \label{fig_overview}
\end{figure*}

\section{Method}
\label{sec:method}

\subsection{Problem Formulation}
 Assume there are $N$ pedestrians in a scene, the spatial location of the $j$-th pedestrian at time $t$ can be denoted as $p_{j,t}=(x_{j,t}, y_{j,t})$. The problem is that given the observed $T_{obs}$ frames as $\{p_{j,t}, j=1,..,N; t=1,...,T_{obs}\}$, we need to predict the trajectories of all pedestrians in the next few frames $\{p_{j,t}, j=1,...,N, t=T_{obs}+1,...,T\}$.

The architecture of our model is depicted in Fig.~\ref{fig_overview}. It consists of three modules including: 1) Encoder: a social graph network for learning both social interactions and individual representation (see Sec.~\ref{sec:method:encoding}); 2) Stochastic: a temporal stochastic model for generating latent variable conditioned on encoder outputs (see Sec.~\ref{sec:method:stochastic}); 3) Decoder: a decoder model to predict the speed of each agent. Given the predicted speed and current location, the next location can be found with a simple addition.

\subsection{Social Graph Network}
\label{sec:method:encoding}

\begin{figure}[h]
\begin{center}
\includegraphics[width=6.0cm]{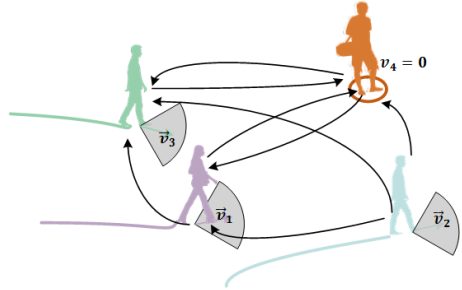}
\end{center}
\caption{An example of social graph.}
\vspace{-3mm}
\label{fig_social}
\end{figure}

At each time $t$, a directed graph $\mathcal{G}_t=(\mathcal{N}_t, \mathcal{E}_t, A_t)$ can be constructed, called social graph. In this graph, each node indicates a pedestrian in the scene, thus the $\mathcal{N}_t$ is not changed throughout the sequence. $\mathcal{E}_t$ represents the set of directed edges determined by adjacency matrix $A_t$. An edge from node $n_i$ to node $n_j$ exists when the element in adjacency matrix ($a_{ij,t}$) equals 1.

As shown in Fig. \ref{fig_social}, we derive a view area for each pedestrian, and construct the timely social graph by inserting edges from all persons inside the view area. For example, two person marked with orange and blue are in the visible view of the person marked with purple. It means the future path of purple person might be affected by these two persons. Thus two edges are added from orange and blue node to the purple node. However, the green person is out of the scope of the purple one, thus no edge exists from purple to green. Specifically, to build the view area, we use the speed direction as the fixation direction of eyes and expand the arc area to a predefined view angle.  In this paper, the view angle is set to $240$ degree, which is larger than maximum human eye angle because of possible eye or head movements. If a person is standing still (as the orange node in Fig. \ref{fig_social}), input edges are inserted from all other nodes. The reason is that the still person can move in any direction and need pay attention to all persons in the scene. Furthermore, a precise view area can be found by estimating head pose if the context image is given. Because the relative positions of pedestrians might be changed during walking, the topology of social graph is not consistent during the whole sequence. At each time, we will update the social graph given the current pedestrians' layout and speed.

Then at time $t$, the embedding representation for each node($j$) and edge($(i,j)$) can be derived as:
\begin{align}
    e_{j,t-1}&= f_{n}([p_{j,t-1}, v_{j,t-1}]) \label{eqn:node_embed}\\
    e_{ij,t-1}&= f_{e}([x_{i,t-1}, x_{j,t-1}, f_{p}(p_{i,t-1}, p_{j,t-1})])\label{eqn:edge_embed}
\end{align}
where $f_{n}, f_{e}$ and $f_{p}$ are neural networks (In this paper, we use one-layer MLP for all $f_{*}$), which encode nodes, edges and pairwise relationships, individually. The input of $f_{p}$ indicates pairwise relationship, which could be measured in two different coordinate systems:
\begin{itemize}
\item Cartesian: $f_{p}(p_{i,t-1}, p_{j,t-1}) = f_{p}(p_{i,t-1} - p_{j,t-1})$. The input is the pairwise position displacement.
\item Polar: $f_{p}(p_{i,t-1}, p_{j,t-1}) = f_{p}(\mathrm{Polar}_{p_{i,t-1}}(p_{j,t-1}))$. The input is the coordinate of $p_{j,t-1}$ in a local polar coordinate system whose reference point is  $p_{i,t-1}$.
\end{itemize}

Experimentally, we found the polar representation performs slightly better than Cartesian representation. The benefits might come from the disentanglement of distance and direction factor.

To obtain social interaction features, a social block is designed to update node representation by accumulating neighborhood information. Formally, at time $t-1$, the update equation of $k$-th social block for $j$-th node is:
\begin{equation}
    x_{j,t-1}^{(k)} = x_{j,t-1}^{(k-1)} + f_s\big(\sum_{\forall i,  a_{ij,t-1}=1} \mathcal{M}_{ij,t-1} x_{ij,t-1}^{(k-1)} \big)
\label{eqn:update}
\end{equation}
where $\mathcal{M}_{ij,t-1}$ denotes the message passing from node $i$ to $j$, and $f_s$ denotes a neural network. Initially, $x_{j,t-1}^{(0)}=e_{j,t-1} $ and $x_{ij,t-1}^{(0)}=e_{ij,t-1}$. $a_{ij,t-1}$ denotes the $ij$ element in the adjacency matrix of the social graph at time $t-1$. In Eqn. (\ref{eqn:update}), the feature of each node will be updated by aggregating information from  its neighborhood nodes.

The message $\mathcal{M}_{ij}$ at time $t-1$ is calculated as:
\begin{equation}
    \mathcal{M}_{ij}^{(k-1)} = \alpha_{ij}^{(k-1)} \cdot ( x_{ij}^{(k-1)}\odot g_{ij}^{(k-1)}).
\label{eqn:message}
\end{equation}

Here we omit subscript $t-1$ for brevity. In equation (\ref{eqn:message}),  $\alpha_{ij}$ is a scalar for edge $(i, j)$, $g$ is the \textit{social gate} for element-wise selection, $\odot$ is the element-wise product operator. Intuitively, the attention value measures the importance of each edge, whereas the social gate acts as element-wise feature selection, similar to the motion gate in ~\cite{Zhang19cvpr}.

We adopt a similar attention calculation as ~\cite{velickovic2018graph}:
\begin{equation}
    \alpha_{ij}^{(k-1)} = \frac{\exp(\mathrm{LeakyReLU}\big(\mathrm{W}^\alpha x_{ij}^{(k-1)})\big)}{\sum_{\forall i, a_{ij}=1} \exp\big(\mathrm{LeakyReLU}(\mathrm{W}^\alpha x_{ij}^{(k-1)})\big)}.
\end{equation}

The social gate is calculated as
\begin{equation}
     g_{ij}^{(k-1)} = \mathrm{sigmoid}\big(f_g(x_{ij}^{(k-1)})\big),
\end{equation}
where $f_g$ is a neural network.

We can sequentially stack multiple (=K ) social blocks. The final output feature of $j$-th pedestrian at time $t-1$ is the output of last social block $x_{j,t-1}^{K}$, which encodes both intrinsic destination and social interactions.


\subsection{Stochastic Trajectory Prediction}
\label{sec:method:stochastic}

In order to generate stochastic predictions, our temporal model samples one each latent variable at each time step. Inspired by ~\cite{Emily18icml}, we define the following update equations:

\begin{align}
p_{\psi}(z_t|x_{<t}) &= \mathrm{LSTM}_{\psi}(x_{t-1}) &  (&prior),\label{prior}\\
q_{\phi}(z_t|x_{\leq t}) &= \mathrm{LSTM}_{\phi}(x_{t}) &  (&inference),\label{inference}\\
p_{\theta}(\hat{v}_t|z_{\leq t},x_{<t}, e_{<t})&= \mathrm{LSTM}_{\theta}(z_t, x_{t-1}, e_{t-1})&  (&generation).\label{generation}
\end{align}
where $x_{t}$ denotes the output node features from social graph network module at time $t$, $z_t$ denotes the sampled stochastic latent variable, $e_{t}$ denotes the node embedding in Eqn. (~\ref{eqn:node_embed}), and $\hat{v}_t$ denotes the output speed prediction. In all three equations, $\mathrm{LSTM}$ models are used to encode past histories.

The prior $\mathrm{LSTM}_\psi$ is learned on its past trajectories with recursive hidden states. The posterior model $\mathrm{LSTM}_\phi$ for inference encodes scenes on the current time step. The prior model is learned to approximate posterior model, in order to capture uncertain social interactions. The detailed descriptions can be found in ~\cite{Emily18icml}.

In generation step, a hierarchical LSTM is used to gradually decode pedestrian features. The first LSTM taking social-encoded features as inputs aim to generate socially-plausible prediction, whereas the second LSTM taking individual embeddings as inputs aim to adjust the predicted path towards individual destination.

Finally, the network is trained end-to-end by maximizing the variational lower bound
\begin{align}
    \mathcal{L}=\sum_{t=1}^T &\Big[\mathbb{E}_{q_{\phi}(z_t|f_{\leq t})}\log p_\theta(y_t|z_{\leq t},f_{<t}) \\
    &-\mathcal{\beta} D_{KL}\big(q_{\phi}(z_t|f_{\leq t})|| p_\psi(z_t|f_{<t})\big)\Big].
\label{eqn:loss}
\end{align}
where the first likelihood term can be reduced to $L_2$ reconstruction loss between the predicted results and ground-truth. The hyper-parameter $\beta$ is chosen as the balance between reconstruction error and sample diversity. We use Gaussian distributions for both the prior and posterior models, and apply reparametrization trick for training with SGD.

\begin{table*}[!htp]
\centering
\small
    \begin{tabular}{|l|c|c|c|c|c|c|}
    \hline
    Method & \multicolumn{6}{c|} {Performance (ADE/FDE)} \\
    \hline
          & ETH & Hotel & Zara01 & Zara02 & Univ & AVG\\
    \hline\hline
    Linear & 1.33/2.94 & 0.39/0.72 & 0.62/1.21 & 0.77/1.48 & 0.82/1.59& 0.79/1.59\\
    LSTM   & 1.14/2.39 & 0.69/1.47 & 0.64/1.43&0.54/1.21 & 0.73/1.60 & 0.75/1.62 \\
    S-LSTM\cite{Alahi16cvpr}& 0.77/1.60&0.38/0.80&0.51/1.19&0.39/0.89&0.58/1.28&0.53/1.15 \\
    SR-LSTM\cite{Zhang19cvpr}*& \textbf{0.63}/\textbf{1.25}&\textbf{0.37}/\textbf{0.74}&0.42/0.90&0.32/0.70&0.51/1.10&\textbf{0.45}/\textbf{0.94}\\
    CVAE\cite{Lee17cvpr}\textdagger&0.93/1.94&0.52/1.03&0.41/0.86&0.33/0.72&0.59/1.27&0.53/1.11\\
    SoPhie\cite{Sadeghian18}*\textdagger&0.90/1.60&0.87/1.82&0.38/0.73&0.38/0.79&0.49/1.19&0.61/1.22\\
    SGAN\cite{Gupta18cvpr}*\textdagger&
    1.19/1.62&1.02/1.37&0.43/0.68&0.58/0.84&0.84/1.52&0.81/1.21\\
    \hline
    Ours\textdagger&0.75/1.63&0.63/1.01&\textbf{0.30}/\textbf{0.65}&\textbf{0.26}/\textbf{0.57}&\textbf{0.48}/\textbf{1.08}&0.48/0.99 \\
    \hline
    \end{tabular}
    \caption{Comparison with other methods. The results marked with * are copied from the paper. The results marked with \textdagger are calculated on multiple $(=20)$ samples.}  \label{tbl:resut_compare}
\begin{tabular}{|lccc|c|c|c|c|c|c|}
    \hline
     \multicolumn{4}{|c|} {Components}&\multicolumn{6}{|c|} {Performance(ADE/FDE)} \\
     \hline
    DG & SG & Polar & K &ETH & Hotel & Zara01 & Zara02 & Univ & AVG \\
     \hline\hline
    $\times$&$\surd$&$\times$&1&0.98/2.01&0.72/1.45&0.51/1.13&0.32/0.71&0.65/1.35&0.64/1.33\\
    $\surd$&$\surd$ &$\times$&1&0.84/1.70&0.66/1.23&0.48/1.13&0.31/0.70&0.60/1.34&0.58/1.22\\
    $\surd$&$\times$&$\times$&2&0.84/1.61&\textbf{0.61}/1.11&0.39/0.88&0.34/0.75&0.67/1.48&0.57/1.27\\
    $\surd$&$\surd$ &$\times$&2&0.81/1.64&0.63/\textbf{1.01}&0.34/0.76&\textbf{0.26}/0.58&0.52/1.17&0.51/1.03\\
    $\surd$&$\surd$ &$\surd$ &2&\textbf{0.75}/\textbf{1.63}&0.64/1.11 &\textbf{0.30}/\textbf{0.65}&\textbf{0.26}/\textbf{0.57}&\textbf{0.48}/\textbf{1.08}&\textbf{0.49}/\textbf{1.01}\\
    \hline
\end{tabular}
    \caption{Evaluation results on different configurations. \textbf{DG} denotes our social directed graph. \textbf{SG} denotes the social gate. \textbf{Polar} denotes polar coordinates. \textbf{K} denotes the number of social blocks.}\label{tbl:component}
\end{table*}

\section{Experiments}
\label{sec:exp}
\subsection{Datasets and Metrics}
\textbf{Datasets:} We evaluated our method on two public datasets: ETH ~\cite{Pellegrini10eccv} and UCY ~\cite{Lerner07ucy}, which consist of rich real-world human-human interactions. These two datasets contain 5 scenes, including 2 scenes (university and hotel) from ETH and 2 scenes (zara and university) from UCY. The average pedestrian number of a scene is 18.0 for UCY and 5.9 for ETH. All the trajectory coordinates are converted to world coordinates and interpolated to sample the coordinate at every 0.4 seconds. In total, there are 1536 pedestrians covering complex social interactions. Following prior work ~\cite{Alahi16cvpr,Zhang19cvpr}, we use the leave-one-out strategy for evaluation. Also we take 8 frames (=3.2 seconds) as observation, and predict the next 12 time steps(=4.8 seconds).

\textbf{Metrics:} Following ~\cite{Alahi16cvpr,Zhang19cvpr}, we evaluate with two error metrics in meters.
\begin{itemize}
   \item Average Displacement Error (ADE): Averaged Euclidean distance between ground-truth and predicted coordinates over all predicted time steps.
   \item Final Displacement Error (FDE): Euclidean distance between ground-truth and predicted coordinates at the last frame.
\end{itemize}

\subsection{Implementation Details}
The dimension of both embedding ($e_j$) and social ($x_j$) features is set to $32$. The dimension of hidden states of is $32$ for both prior and posterior LSTMs, and $64$ for decoder LSTMs. Table \ref{tbl:detailed_network} details  network configuration with one social block. The batch size is $16$ scenes with variable pedestrian number. The Adam optimizer is adopted with an initial learning rate of $5e-4$, the epoch number is 300, and $\beta=1e-4$ for all experiments.

During prediction, we use the one-step mode, indicating that we iteratively use the previous prediction results as the inputs of the step. In contrast, during training, the inputs are always the ground-truth of the last frame.

\subsection{Comparison with Existing Methods}
\label{sec:exp:compare}

\textbf{Baselines:} A few baselines are used for comparison, including both deterministic and stochastic methods. For deterministic methods, we choose
\textit{Linear} (a linear regressor trained by minimizing least square error) and three LSTM-based methods, including vanilla LSTM (denoted as \textit{LSTM}), social LSTM (denoted as \textit{S-LSTM}) and \textit{SR-LSTM}. For stochastic methods, we choose \textit{CVAE} (we use the same network settings as ~\cite{Lee17cvpr}) and two methods which introduces stochastic with Gaussian white noise (denoted as \textit{SoPhie}~\cite{Sadeghian18} and \textit{SGAN}~\cite{Gupta18cvpr}). For fair comparison, we consider the model with only trajectory inputs, without the scene images in SoPhie. For stochastic methods, 20 samples are generated for evaluation, whereas the deterministic methods only produce one best prediction.

As shown in Table.~\ref{tbl:resut_compare}, our method can achieve comparable results with the current state-of-the-art methods. In particular, the error reduction is signification in UCY as compared with ETH datasets. Because UCY contains more crowded scenes with complex social interactions, our method demonstrates its superiority when dealing with complicated non-linear trajectories in crowded scenarios, benefited from our social graph network. For ETH, our results are slightly worse than SR-LSTM, but still better than other stochastic methods. Because of simple interactions and few ambiguous paths in ETH, deterministic methods have advantages by optimizing reconstruction loss only.

\begin{figure*}[!ht]
    \begin{minipage}{.99\textwidth}
        \centering
        \begin{tabular}{ccc}
        \includegraphics[width=3.5cm,height=2cm]{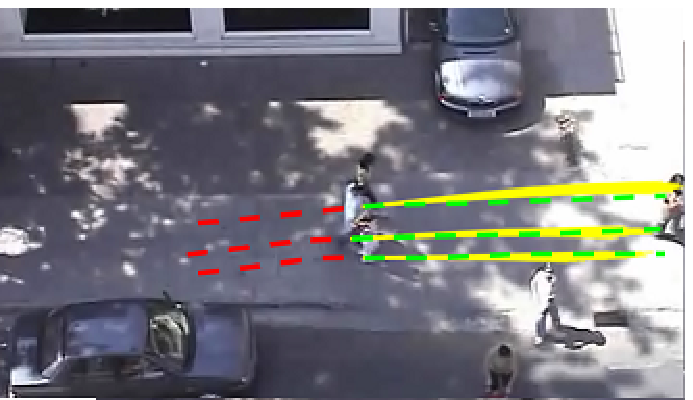}&
        \includegraphics[width=3.5cm,height=2cm]{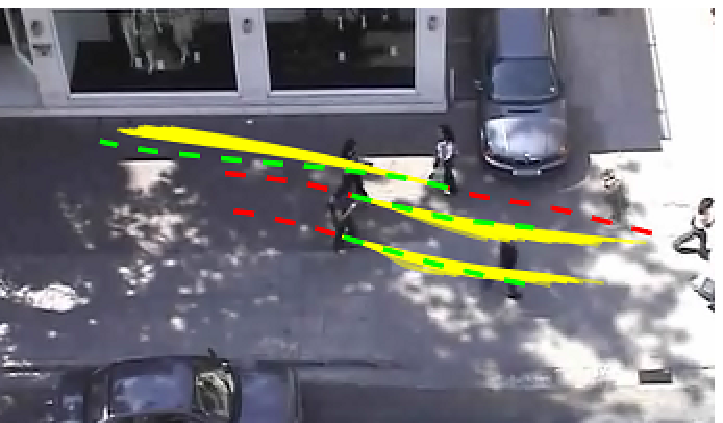}&
        \includegraphics[width=3.5cm,height=2cm]{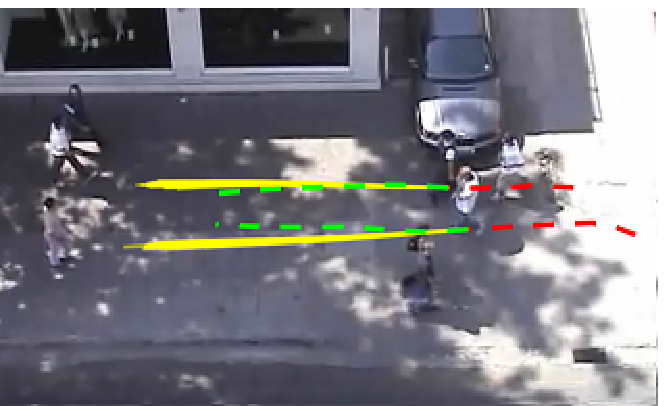}\\
        (a)&(b)&(c)\\
        \includegraphics[width=3.5cm,height=2cm]{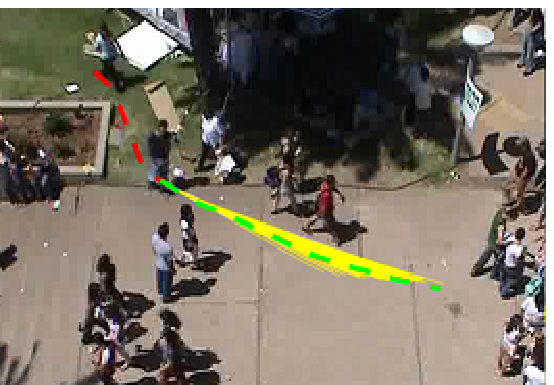}&
        \includegraphics[width=3.5cm,height=2cm]{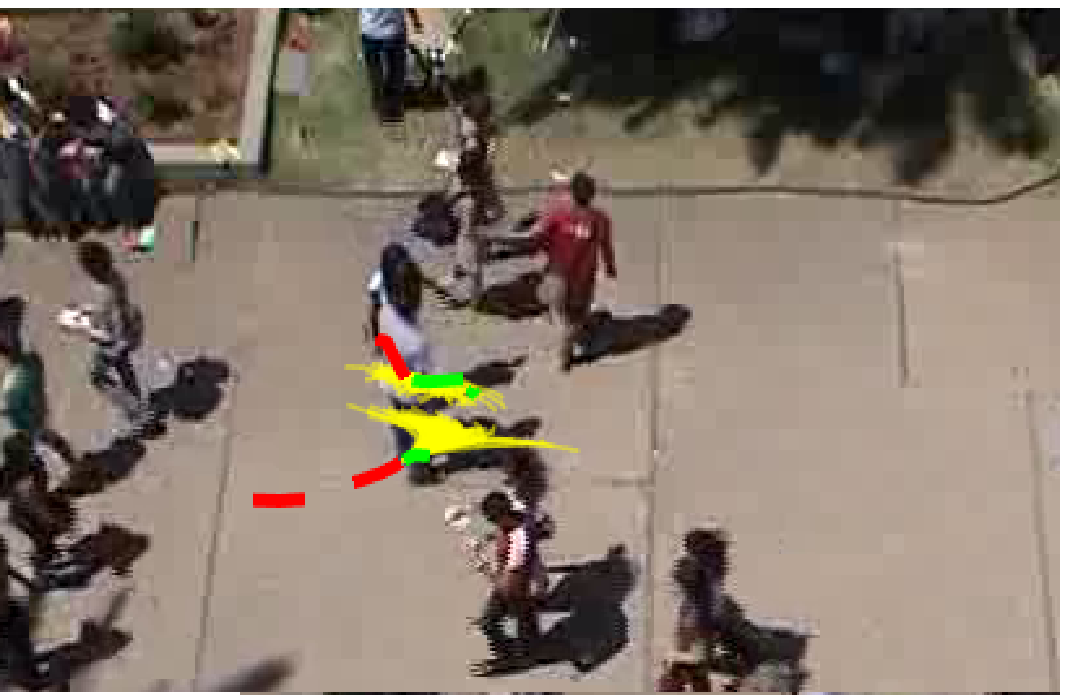}&
        \includegraphics[width=3.5cm,height=2cm]{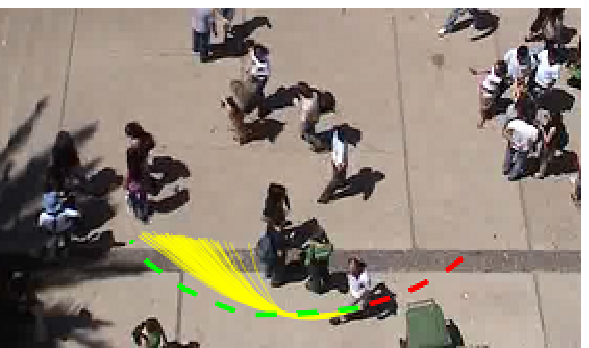}\\
        (d)&(e)&(f)
        \end{tabular}
        \caption{Illustration of the prediction trajectories with various social interactions: (a). Group walking (b). Crossing with group. (c). Person following. (d). Person crossing. (e). Person merging. (f). Group avoidance. Here {\color{red} \textbf{red line}}: observed history {\color{green} \textbf{green line}}: future ground-truth {\color{yellow} \textbf{yellow line}}: our multiple prediction results. (Best view in color)}\label{fig_traj_viz}
    \end{minipage}\vfill
    \begin{minipage}{.99\textwidth}
        \centering
        \begin{tabular}{ccc}
        \includegraphics[width=3.5cm,height=2cm]{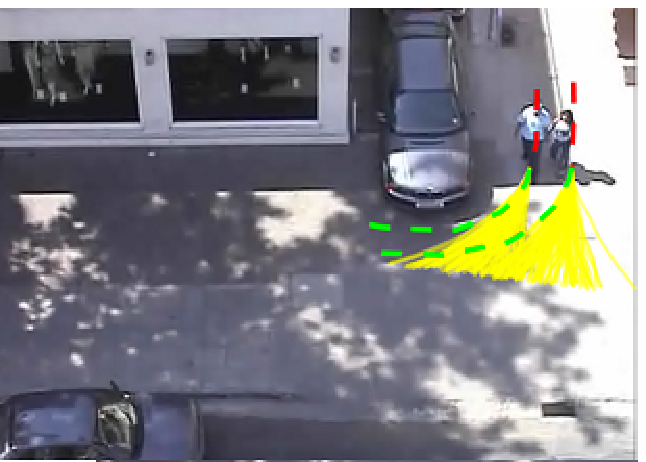} &
        \includegraphics[width=3.5cm,height=2cm]{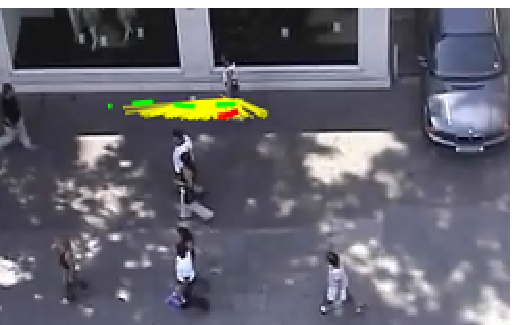} &
        \includegraphics[width=3.5cm,height=2cm]{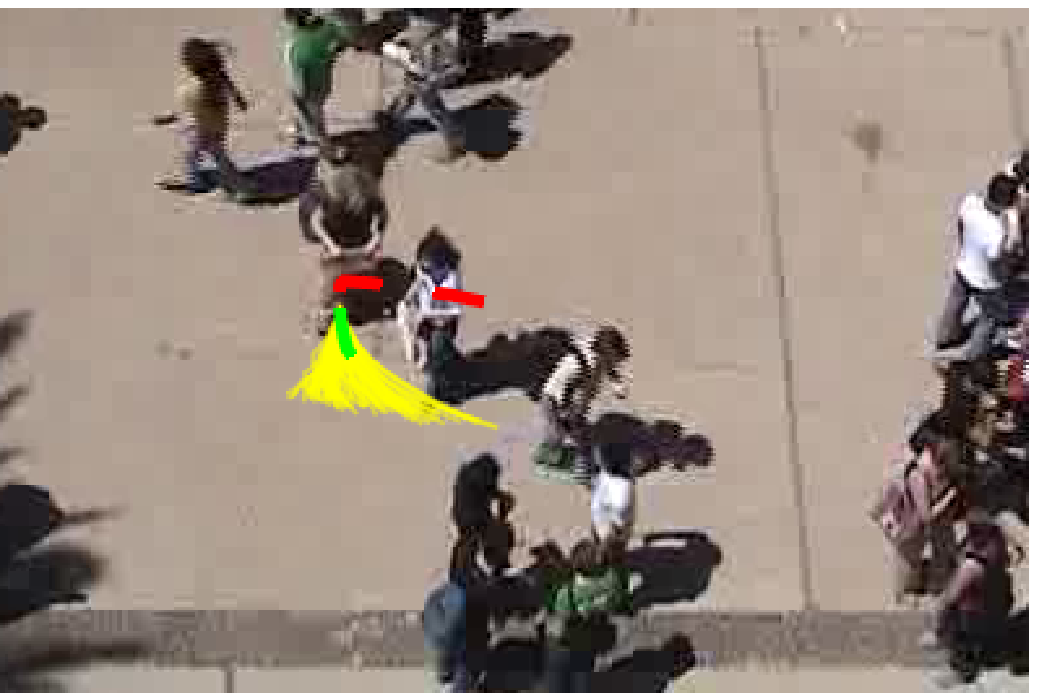}\\
        (a)&(b)&(c)
        \end{tabular}
        \caption{Example of diverse predictions from our model. (a) Cross road (b). Standing person (c). Sudden turn.}\label{fig_stochastic}
    \end{minipage}\vfill
    \begin{minipage}{.99\textwidth}
        \centering
        \begin{tabular}{ccc}
        \fbox{\includegraphics[width=3.5cm,height=2cm]{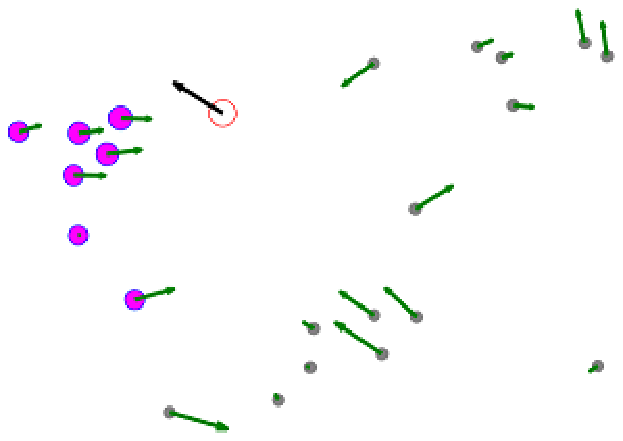}}&
        \fbox{\includegraphics[width=3.5cm,height=2cm]{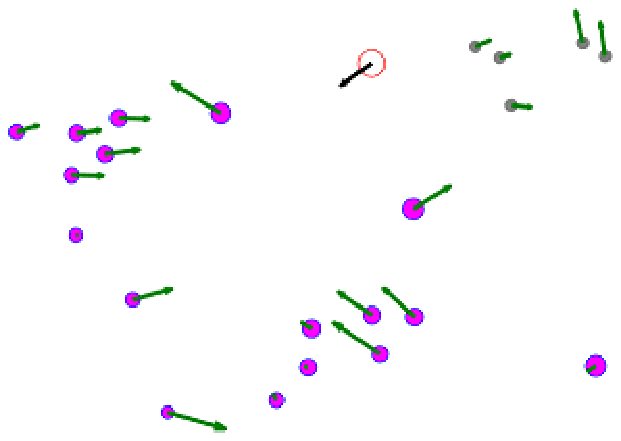}}&
        \fbox{\includegraphics[width=3.5cm,height=2cm]{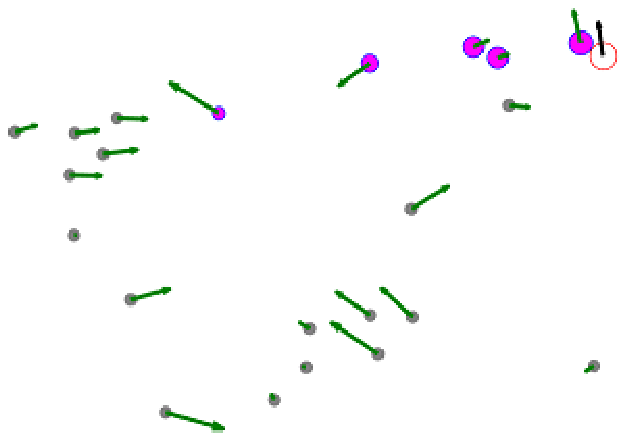}}\\
        \end{tabular}
        \caption{Illustration of attention values for different target pedestrian under the same scene. Each circle represents a pedestrian in the scene. Circle in red is the target pedestrian, who pays attention to pedestrians denoted by magenta nodes and no attention to gray nodes. The black arrow gives the velocity of the target, whereas the green arrow gives the velocity of other pedestrians. The radius of the magenta circle is in proportion to attention value.} \label{fig_attention}
    \end{minipage}
\end{figure*}

\subsection{Ablation Study}
\subsubsection{Component Analysis}

Table \ref{tbl:component} gives results on several configurations of our method varied in whether to use our directed graph, social gate, polar coordinates and different number of blocks in social graph network. When the directed graph is disabled, an undirected fully-connected graph topology is used, in which all elements in the adjacency matrix are 1. From the table, it is worth noting that the directed social graph can significantly reduce the error from $0.64$ to $0.58$, which indicates that the selection of noticeable pedestrians is important for boosting performance. Another salient error reduction comes from the introduction of social gate, which indicates element-wise social feature selection helps to filter information during message passing. In general, more refinements ($K=2$) perform better than one step social calculation.

\subsubsection{Qualitative Analysis}

\textbf{Social aware prediction}. In Fig. \ref{fig_traj_viz}, we illustrate six crowd scenarios where the target person has to adjust his path towards the destination. As shown, our method can learn social norms and have the ability to adjust the path towards destination. For example, when meeting with a group, as shown in Fig. \ref{fig_traj_viz} (b) and (f), our prediction make a detour in order to avoid stepping into the group. In Fig. \ref{fig_traj_viz} (d), our results give reasonable routines to walk through the crowds without collision. Our method can also capture potential social intention, such as generating a new group in Fig.~\ref{fig_traj_viz} (e). More results can be found in the demo videos.

\textbf{Stochastic movement}. Generally, the prediction is uncertain, especially when walking at a low speed or near the road crossing. Fig.~\ref{fig_stochastic} shows our prediction results in these cases. As shown in Fig.~\ref{fig_stochastic}, in the road corner, our method gives two options: go straight or turn right. It is worth noting that the generated stochastic predictions still do not break the consistency group walking.

\textbf{Social attention}. Fig.~\ref{fig_attention} illustrates the attention value of some example pedestrians in the same crowded scene. It shows that our directed graph can help to filter irrelevant pedestrians (marked in gray circles). The dominant attention is paid to the neighboring person, while he still notices other pedestrians which might affect his routine.

\section{Conclusions}
In this paper, we propose a temporal stochastic model with social graph network to address the problem of predicting all social plausible trajectories in the crowds. We propose a directed social graph and a network to encode both individual and social features. In addition, we utilize a temporal stochastic model which sequentially learns dynamic prior model at each time step. The final one-step prediction is generated by sampling from the prior model and progressively decoded with a hierarchical LSTMs. Our empirical evaluations on real data-sets demonstrate our improvement over the current state-of-the-art methods in crowded scenes. In the future, we plan to introduce context images to refine our social graph construction and add scene semantics, such as obstacles and road path, derived from context images.

\begin{table*}
    \centering
    \begin{tabular}{l|l|l|l}
    \hline
       Layer & Input, (Dimensions) &Output, (Dimensions) & Parameters\\
    \hline
        \multicolumn{4}{c} {Encoder}\\
    \hline
        Fully-connected & $[p_j,v_j]$, (4)  & $f_{n,j}$, (32) &act:=ReLU\\
        Fully-connected & $\mathrm{Polar}_{p_i}[p_j,v_j]$, (4) & $f_{p,ij}$, (32)&act:=ReLU\\
        Fully-connected & $[f_{n,i}, f_{n,j}, f_{p,ij}]$, (96)& $f_{e,ij}$, (32) &act:=ReLU\\
        Fully-connected & $[f_{n,i}, f_{n,j}, f_{p,ij}]$, (96)& $W^\alpha_{ij}$, (32) &act:=LeakyReLU\\
        Softmax   & $W^\alpha_{ij}$, (32) & $\alpha_{ij}$, (1) &\\
        Fully-connected & $[f_{n,i}, f_{n,j}, f_{p,ij}]$, (96)& $f_{g,ij}$, (32) &act:=ReLU \\
        Sigmoid & $f_{g,ij}$, (32) & $gate_{ij}$, (32) &\\
        Identity & $f_{n,j}$, (32)& $x_{j}^{(0)}$, (32) & $x_{j}^{(0)}=f_{n,j}$\\
        Identity & $f_{e,ij}$, (32)& $x_{ij}^{(0)}$, (32) &$x_{ij}^{(0)}=f_{e,ij}$\\
        Multiplication& $x_{ij}^{(0)},\alpha_{ij}, gate_{ij}$, (32),(1),(32) & $\mathcal{M}_{ij}$,(32) & $\mathcal{M}_{ij}=\alpha_{ij}\cdot(x_{ij}^{(0)}\odot gate_{ij})$\\
        Aggregation& $\mathcal{M}_{ij}, x_{ij}^{(0)}$, (32),(32) & $\mathcal{M}x_{j}^{(0)}$, (32) & $\mathcal{M}x_{j}^{(0)} =\sum_{i,\forall a_{ij}=1}\mathcal{M}_{ij} x_{ij}^{(0)}$\\
        Fully-connected & $\mathcal{M}x_{j}^{(0)}$, (32)& $f_{s,j}$,(32) & \\
        Addition& $x_{j}^{(0)}$, $f_{s,j}$, (32),(32) & $x_{j}^{(1)}$, (32) &$ x_{j}^{(1)} =x_{j}^{(0)} +f_{s,j}$\\
     \hline
        \multicolumn{4}{c} {Prior}\\
    \hline
        Fully-connected & $x_{j,t-1}^{(1)}$, (32)& $f_{prior}(j,t-1)$, (32)&\\
        LSTM  & $f_{prior}(j,t-1)$, (32)& $h_{prior}(j,t-1)$, (32)&\\
        Fully-connected & $h_{prior}(j,t-1)$, (32)& $\mu_j$, (32) & \\
        Fully-connected & $h_{prior}(j, t-1)$, (32) & $\sigma_j$, (32)& act:=$\frac{1}{2} exp(\dot)$\\
        Reparam. trick &$\mu_j, \sigma_j$, (32),(32)& $z_j$, (32)& $z_j=\mu_j+\sigma_j\odot \epsilon_j, \epsilon_j\sim \mathcal{N}(0,1)$\\
    \hline
        \multicolumn{4}{c}{Inference}\\
    \hline
        Fully-connected & $x_{j,t}^{(1)}$, (32)& $f_{inf}(j,t)$, (32)&\\
        LSTM  & $f_{inf}(j,t)$, (32)& $h_{inf}(j,t)$, (32)&\\
        Fully-connected & $h_{inf}(j,t)$, (32)& $\mu_j^p$, (32) & \\
        Fully-connected & $h_{inf}(j,t)$, (32) & $\sigma_j^p$, (32)&act:=$\frac{1}{2} exp(\dot)$ \\
    \hline
        \multicolumn{4}{c} {Decoder}\\
    \hline
        LSTM  & $[x_j^{(1)}, z_j]$, (64) & $h_j^1$, (32)&\\
        LSTM &$[h_j^1, f_{n,j}]$, (64) & $h_j^2$, (64) & \\
        Fully-connected & $h_j^2$, (64)&$\hat{v}_j$, (2)& \\
    \hline

    \hline
    \end{tabular}
    \caption{Detailed architecture of our network with one social block.}
    \label{tbl:detailed_network}
\end{table*}

{\small
\bibliographystyle{ieee_fullname}
\bibliography{egbib}
}
\end{document}